\documentclass{article}
\usepackage{spconf,amsmath,graphicx}
\usepackage{amssymb}

\usepackage{enumitem}
\usepackage{booktabs}
\usepackage{siunitx}
\usepackage{caption}
\usepackage{threeparttable}
\setlist{nosep, leftmargin=14pt}

\usepackage{mwe} 

\title{Large-scale EM Benchmark for Multi-Organelle Instance Segmentation in the Wild}

\name{
    {\em Yanrui Lu$^{1,2}$ \quad
    Danyang Chen$^{1}$ \quad
    Haowen Xiao$^{1}$ \quad
    Jiarui Zhu$^{1}$ \quad
    Fukang Ge$^{1}$ \quad
    Binqian Zou$^{1}$} \\
    {\em Jiali Guan$^{1}$ \quad
    Jiayin Liang$^{1}$ \quad
    Yuting Wang$^{1}$ \quad
    Ziqian Guan$^{1}$ \quad
    Xiangcheng Bao$^{1}$ \quad
    Jinhao Bi$^{3}$} \\
    {\em Lin Gu$^{4}$ \quad
    Jun He$^{1}$ \quad
    Yingying Zhu$^{1,*}$
    \thanks{$*$ Corresponding author.}}
}

\address{
    $^{1}$ Guangzhou Institutes of Biomedicine and Health, Chinese Academy of Sciences, China \\
    $^{2}$ South China University of Technology, China \\
    $^{3}$ Center for Infectious Disease Research, Westlake University, China \\
    $^{4}$ RIKEN, Japan
}

\usepackage{color}
\begin{document}
%
\maketitle

%



\begin{abstract}
Accurate instance-level segmentation of organelles in electron microscopy (EM) is critical for quantitative analysis of subcellular morphology and inter-organelle interactions. However, current benchmarks, based on small, curated datasets, fail to capture the inherent heterogeneity and large spatial context of in-the-wild EM data, imposing fundamental limitations on current patch-based methods.
To address these limitations, we developed a large-scale, multi-source benchmark for multi-organelle instance segmentation, comprising over 100,000 2D EM images across variety cell types and five organelle classes that capture real-world variability. Dataset annotations were generated by our designed connectivity-aware Label Propagation Algorithm (3D LPA) with expert refinement. We further benchmarked several state-of-the-art models, including U-Net, SAM variants, and Mask2Former.
Our results show several limitations: current models struggle to generalize across heterogeneous EM data and perform poorly on organelles with global, distributed morphologies (e.g., Endoplasmic Reticulum). These findings underscore the fundamental mismatch between local-context models and the challenge of modeling long-range structural continuity in the presence of real-world variability. The benchmark dataset and labeling tool will be publicly released soon.
\end{abstract}

\begin{keywords}
Instance-level Segmentation, Electron Microscopy, Organelle Segmentation
\end{keywords}

\section{Introduction}
\label{sec:intro}

\begin{figure}[!t] 
\begin{minipage}[b]{1.0\linewidth}
  \centering
  \centerline{\includegraphics[width=1.1\linewidth]{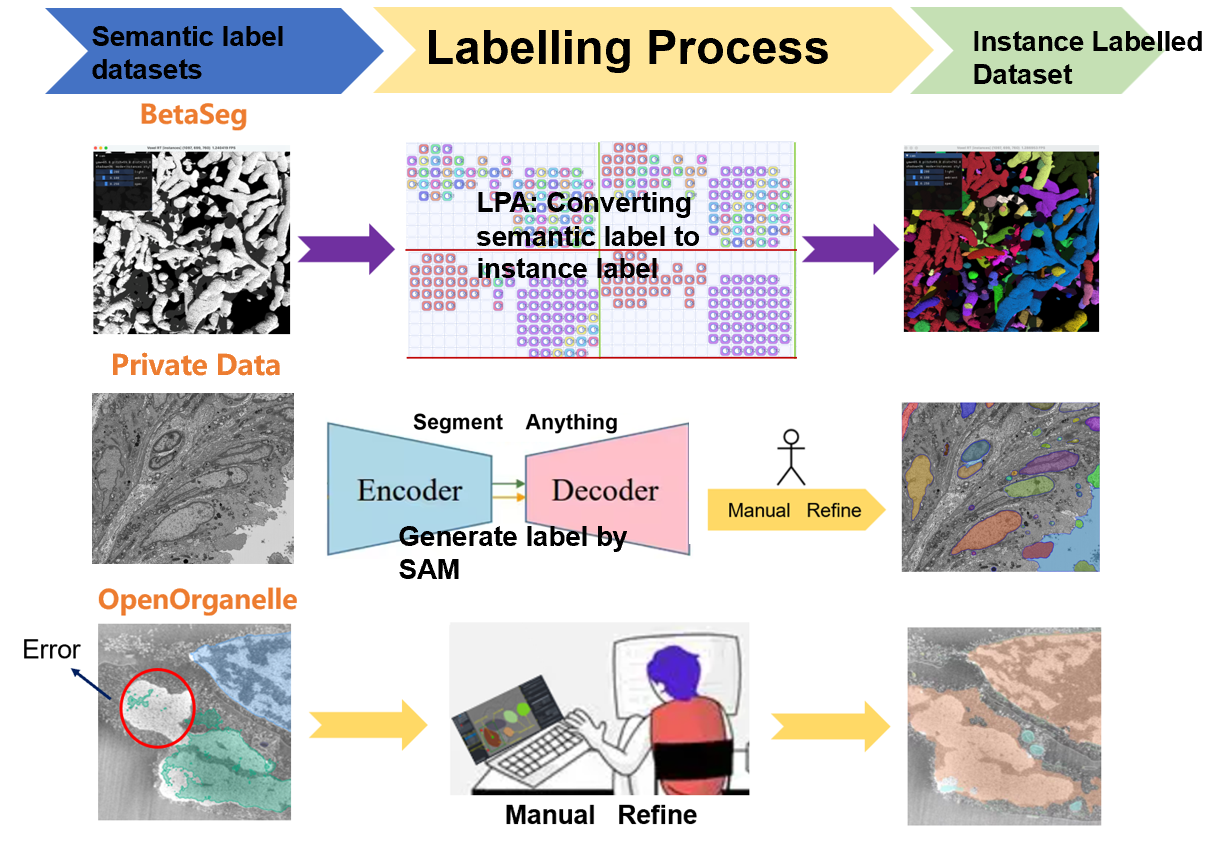}}
\end{minipage}
\caption{Overview of the mixed label processing pipeline.}
\label{fig:1}
\end{figure}

\begin{figure}[!t] 
\begin{minipage}[b]{1.0\linewidth}
  \centering
  \centerline{\includegraphics[width=1.1\linewidth]{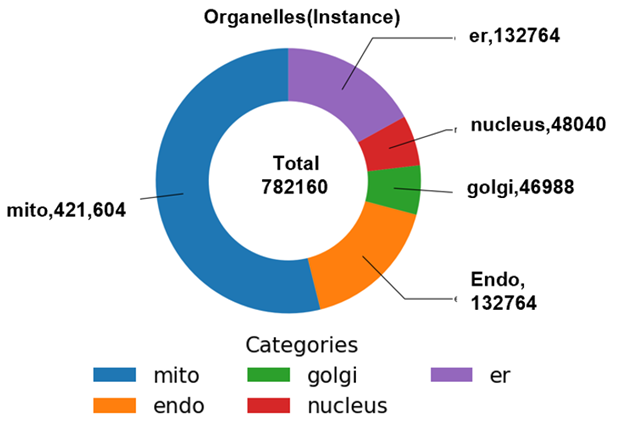}}
\end{minipage}
\caption{Composition of the proposed benchmark across data.}
\label{fig:2}
\end{figure}

Electron microscopy (EM) provides nanometer-scale resolution essential for analyzing subcellular morphology and organelle interactions. Segmentation is key for EM studies and biophysical modeling. Early efforts, such as the Lucchi dataset\cite{Lucchi2012TMI}, applied supervoxels and U-Net-style CNNs for semantic mitochondria segmentation. Lucchi++\cite{Kasthuri2015Cell} improved boundary quality, while MitoEM\cite{Wei2020MitoEM} enabled 3D instance segmentation on large FIB-SEM volumes. OpenOrganelle\cite{Heinrich2021Nature} extended this to whole-cell multiclass tasks. However, these benchmarks often rely on small, curated crops or homogeneous datasets, thereby avoiding the challenges of heterogeneity and large-scale context present in real-world data.
Recent methods, including MitoNet\cite{Conrad2022.03.17.484806}, DualRel\cite{mai2023dualrel}, and micro-SAM\cite{archit2025segment}, improved generalization and feature extraction. Still, most public datasets focus on single organelles and semantic segmentation. current models remain limited in end-to-end multi-organelle instance segmentation.

Instance-level segmentation is critical for quantifying organelle morphology, spatial organization, and inter-organelle topology. This granularity is essential for downstream applications such as tracking dynamic changes, modeling physical interactions, and identifying organelle-specific phenotypes at the systems level.

Despite its scientific importance, the field remains trapped in semantic segmentation due to a fundamental mismatch between large-scale, global 'in-the-wild' EM data and SOTA models optimized for local patches. Standard "croping mechanisms" fail to bridge this gap: curated crops avoid scale challenges, tiling fragments global instances, and resizing distorts native resolution.

To break this impasse, we make two contributions. First, to overcome the annotation barrier, we develop a hybrid pipeline using 3D Connected Component Labeling (3D CCL) and expert refinement (Sec 2.2). Second, we use this benchmark to quantify the critical failure of patch-based models on global structures. Our key contributions are:
\begin{itemize}
\item We present a new large-scale, "in-the-wild" benchmark for instance-level multi-organelle segmentation, built using our hybrid 3D LPA and expert-refinement pipeline.
\item We benchmark representative SOTA methods and quantify the fundamental mismatch between patch-based methods and "in-the-wild" data.
\item We demonstrate that this mismatch causes models to succeed on local organelles (Mito, Nucleus) but catastrophically fail on global network organelles (ER, Golgi), revealing a critical bottleneck for the field. We will release our dataset and tool upon acceptance.
\end{itemize}

\section{Benchmark}
\label{sec:benchmark}

\subsection{Data Acquisition}
\label{ssec:data_acquisition}

We have drawn over 100 thousands 2D EM images from open source data (OpenOrganelle, BetaSeg\cite{muller20203d}) and combined our private data which have already applied stitching and registration to it. These data span over variety cell types and five organelle types. In order to make our model to have the ability to segment in wild which require a Benchmark reflecting real-world scenarios, we kept images with dynamic resolution and multi-scale information.

\subsection{Data Refine and Annotation}
\label{ssec:data_annot}

We propose a mixed label processing procedure to unify heterogeneous datasets in order to build a multi-organelle instance segmentation task, shown in Fig.\ref{fig:1}. Since most of the existing works are annotated by a semi-automatic workflow which would inevitably introduce numerous errors, we refined them by organizing an expert group consisting of one expert and six students. For dataset like BetaSeg\cite{muller20203d} which only provides semantic labels, we proposed a 3D LPA(3D Connected Component Labeling) algorithm (detailed in Sec 3.2) to convert semantic label into instance label. At last, we still facilitated a semi-automatic approach to annotate the dataset we provide. We used models pre-trained on other open-source datasets to gain coarse-grained label and asked the expert group to proofread until no disagreement. And the composition of the benchmark is shown in Fig.\ref{fig:2}.

\subsection{Data Composition and Analysis}
\label{ssec:data_analysis}

We evaluated the initial output of our 3D LPA tool (Sec 3.2) prior to expert refinement and found it achieved a high baseline accuracy. For example, on the BetaSeg dataset, an estimated 73 percent of instances were correctly generated automatically. The remaining cases, which were the focus of our expert refinement (Sec 2.2).

\begin{figure}[!t] 
\begin{minipage}[b]{1.0\linewidth}
  \centering
  \centerline{\includegraphics[width=1.1\linewidth]{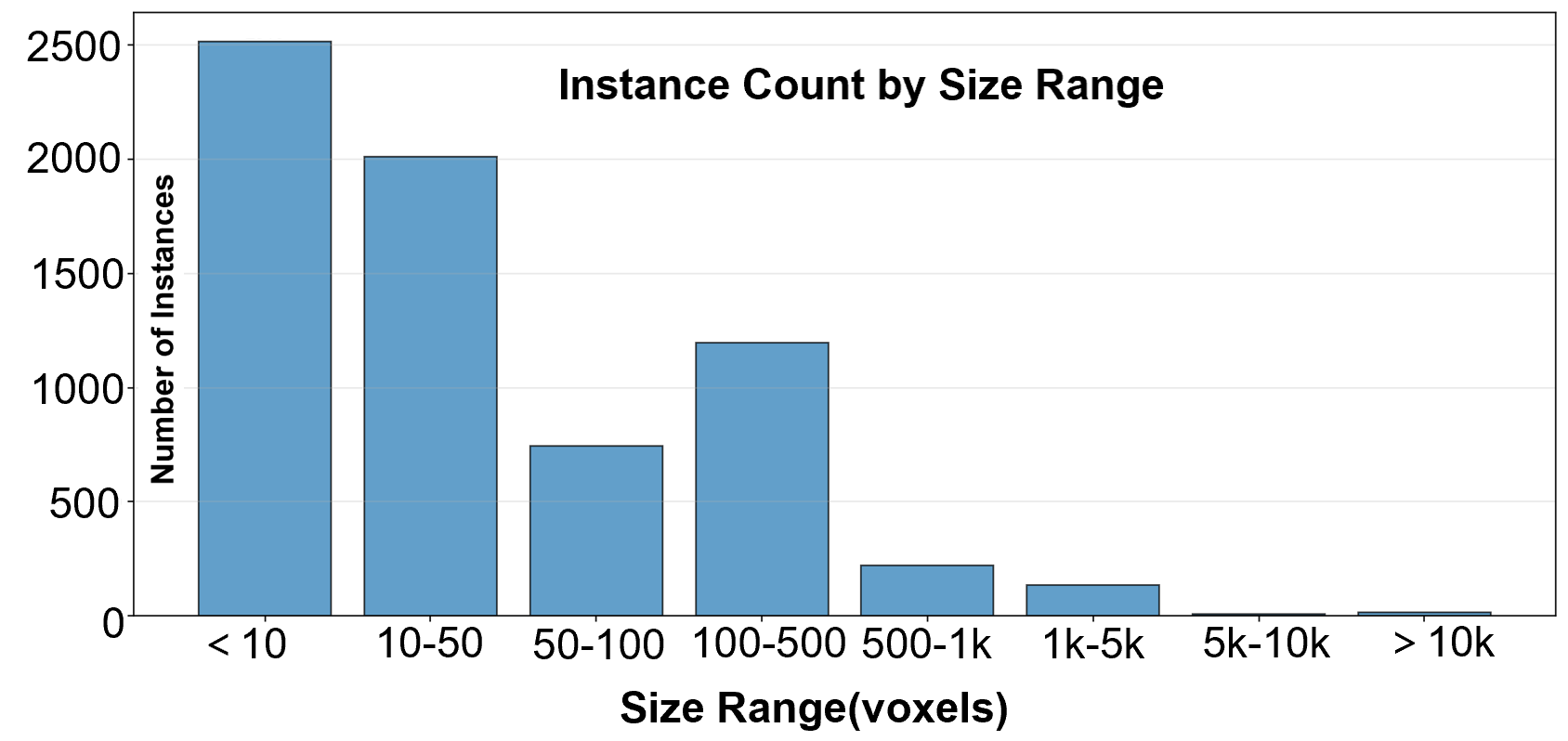}}
\end{minipage}
\caption{Distribution of mitochondria instance sizes in the converted BetaSeg subset.}
\label{fig:3}
\end{figure}

\begin{figure}[!t] 
\begin{minipage}[b]{1.0\linewidth}
  \centering
  \centerline{\includegraphics[width=\linewidth]{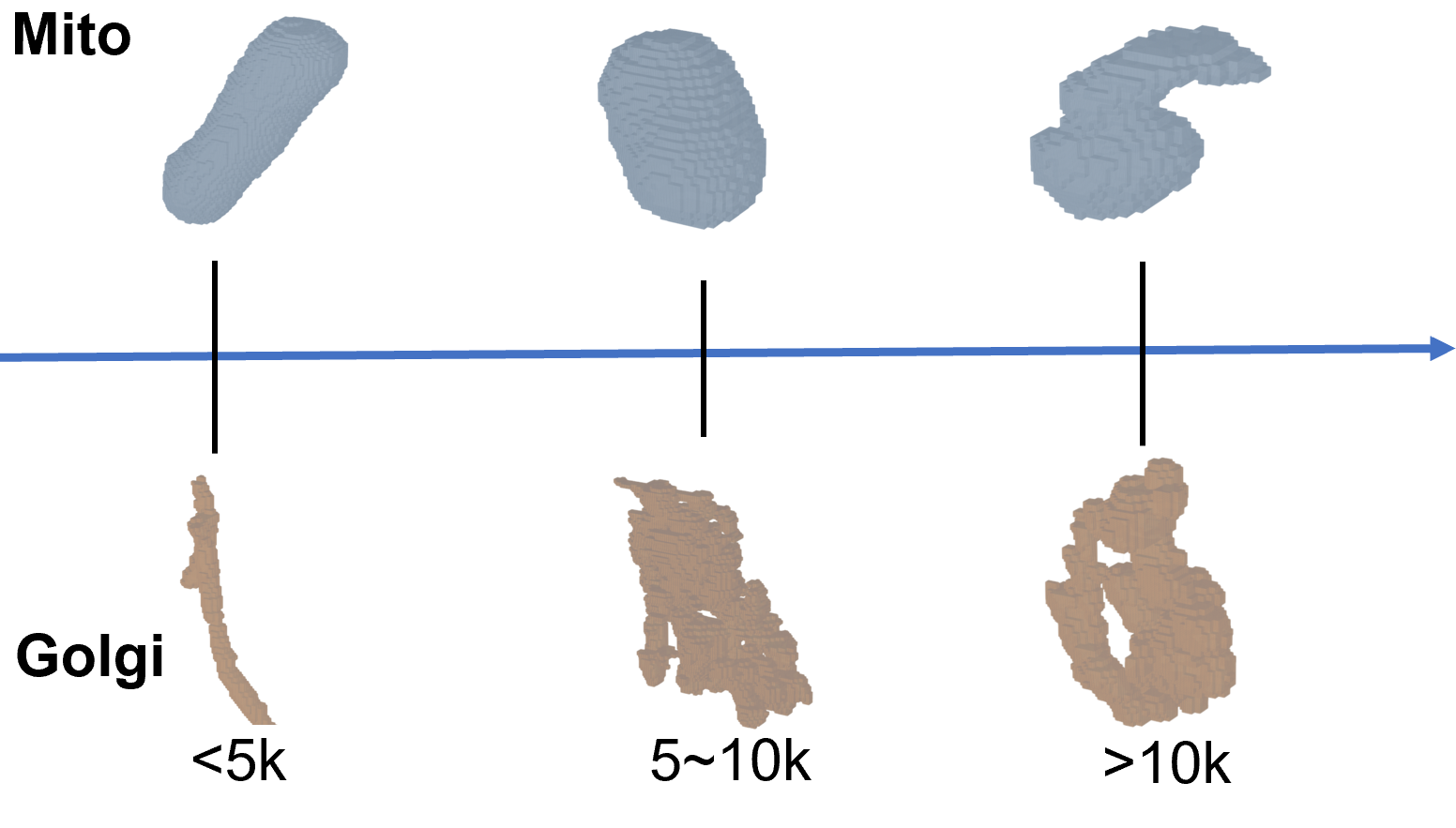}}
\end{minipage}
\caption{Representative 3D meshes of mitochondria and Golgi instances with small, medium, and large volumes.}
\label{fig:4}
\end{figure}

After this refinement process, we analyzed the final dataset composition. We figured that the converted BetaSeg dataset alone had over 120k instances. Take mitochondria as example, we show its distribution of instance sizes in Fig.\ref{fig:3}. To illustrate the diverse morphology of organelle, we pick some 3D mitochondria and golgi samples from small size ($<$5k voxels), middle size(5k-10k) to large size ($>$10k voxels), shown in Fig.\ref{fig:4}.

\section{Method}
\label{sec:format}

Our overall method consists of a new large-scale benchmark and a novel labeling tool. To validate this benchmark and probe its challenges, we define a rigorous benchmarking methodology that utilizes three representative state-of-the-art architectures as baseline probes.

\subsection{Segmentation model}
\label{ssec:seg_model}

To establish strong baselines for our benchmark, we evaluated three representative paradigms in modern instance segmentation: a classic bottom-up pipeline (U-Net), an end-to-end query-based transformer (Mask2Former), and a prompt-based foundation model (SAM).

\noindent\textbf{U-net}\cite{ronneberger2015u}:
We employ U-Net as a strong semantic baseline for both 2D and 3D EM. The network uses an encoder–decoder with skip connections to fuse high-resolution localization with deep semantic context. The network predicts per-pixel class logits for all organelle categories and, optionally, an auxiliary boundary channel to sharpen interfaces between touching objects. Instance masks are decoded from the semantic output using connected-components and marker-controlled watershed on distance (or affinity) maps, a standard practice in EM instance segmentation. 

\noindent\textbf{SAM(Segmentation Anything)}:
SAM is a vision transformer–based, promptable segmenter comprising a powerful image encoder and a lightweight mask decoder conditioned on points/boxes/text prompts. Instead of training or finetuning, we used two weights, which are vanilla SAM and micro-sam, to evaluate on our benchmark. In the process we reported two settings: (i) automatic segmentation using SAM’s mask generator to emulate fully automatic use (ii) prompt-guided segmentation useed synthetic prompts derived from ground-truth: positive points from object centroids, negative points near boundaries and ground-truth bounding boxes.

\noindent\textbf{Mask2Former}\cite{cheng2022masked}:
Mask2Former formulates segmentation as predicting a set of masks and their categories via transformer queries with multi-scale deformable attention. A shared decoder iteratively updates queries that attend to image features and output class scores and mask embeddings; bipartite matching (Hungarian assignment) aligns predictions to ground-truth instances. We adopted the standard instance-segmentation objective and treated each organelle type as a distinct category. Compared with bottom-up decoding, Mask2Former provides direct instance predictions, offering a complementary perspective on multi-organelle segmentation quality.

\subsection{Label converting tool}
\label{ssec:label_tool}

We proposed a Label Propagation Algorithm, 3D LPA to convert semantic label into high-quality instance label. We referenced connected component labeling algorithms from the OpenCV\cite{opencv-python} library and improved upon them to develop a 3D version.
Let $V \subset \mathbb{Z}^3$ represent a discrete 3D voxel space. Let $S: V \to \mathbb{Z}_{\geq 0}$ be the input semantic volume, where $S(p)$ is the semantic class label of voxel $p \in V$. By convention, $S(p)=0$ indicates background. We identify connected components by iteratively updating an intermediate label volume $L$. At iteration $t=0$ , assign a unique initial label to each foreground voxel. Use a function $\text{UID}(p)$ that maps a voxel coordinate to a unique positive integer (e.g., the voxel’s linear index plus 1, as in the code):

$$
L^{(0)}(p) =
\begin{cases}
\text{UID}(p) & \text{if } S(p) > 0 \\
0 & \text{else}
\end{cases}
$$

At each iteration $t+1$, every voxel $p$ updates its label to the minimum among its own label and those of all adjacent voxels. Let $\mathcal{A}(p) = \{q \in V \mid p \sim q\}$ be the adjacency set of $p$ . The update rule is:

\[
\resizebox{0.99\linewidth}{!}{$
L^{(t+1)}(p)=
\begin{cases}
\min\bigl(\{L^{(t)}(p)\}\cup\{L^{(t)}(q)\mid q\in\mathcal{A}(p)\}\bigr) & \text{if } S(p)>0,\\
0 & \text{else}
\end{cases}
$}
\]

The algorithm will continue iterate until convergence, $L^{(T)} = L^{(T-1)}$.At convergence, all voxels in the same connected component share the minimal initial UID within that component (the “representative label”).

\begin{figure}[!t] 
\begin{minipage}[b]{1.0\linewidth}
  \centering
  \centerline{\includegraphics[width=\linewidth]{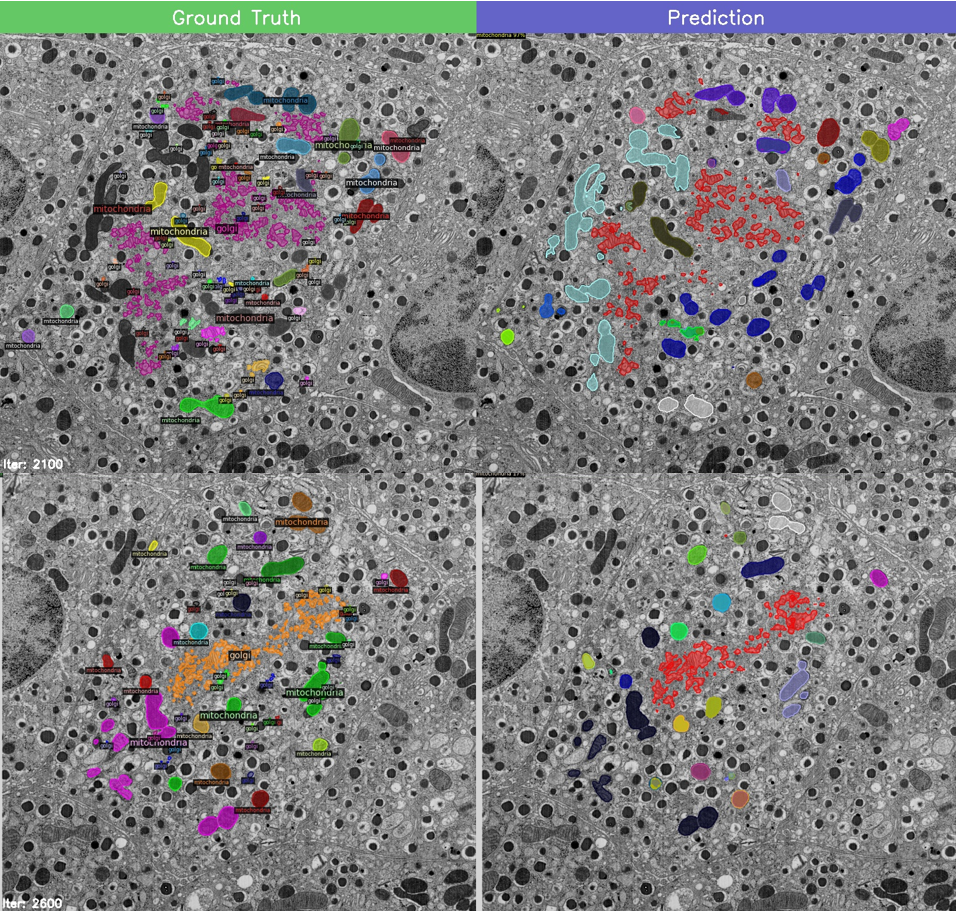}}
\end{minipage}
\caption{Qualitative visualization of multi-organelle instance labels, where each color denotes a distinct instance.}
\label{fig:res}
\end{figure}

\sisetup{
  table-number-alignment = center,
  table-figures-integer = 1,
  table-figures-decimal = 3,
  detect-weight=true,
  detect-inline-weight=math
}

\begin{table}[t]
  \centering
  \begin{threeparttable}
\setlength{\tabcolsep}{3pt}
  \begin{tabular}{
    l
    S[table-format=1.3] S[table-format=1.3]
    S[table-format=1.3] S[table-format=1.3]
    S[table-format=1.3] S[table-format=1.3]
    S[table-format=1.3] S[table-format=1.3]
    S[table-format=1.3] S[table-format=1.3]
  }
  \toprule
  \multicolumn{1}{c}{\textbf{}} &
  \multicolumn{2}{c}{\textbf{Mito}} &
  \multicolumn{2}{c}{\textbf{Nucleus}} &
  \multicolumn{2}{c}{\textbf{ER}} &
\\
  \cmidrule(lr){2-3}
  \cmidrule(lr){4-5}
  \cmidrule(lr){6-7}

  & \multicolumn{1}{c}{Dice$\uparrow$} & \multicolumn{1}{c}{IoU$\uparrow$}
  & \multicolumn{1}{c}{Dice$\uparrow$} & \multicolumn{1}{c}{IoU$\uparrow$}
  & \multicolumn{1}{c}{Dice$\uparrow$} & \multicolumn{1}{c}{IoU$\uparrow$}
 \\
  \midrule
  
  Unet         & 0.692 & 0.603 & 0.879 & 0.916 & 0.248 & 0.158 &  \\
  SAM(a)       &   x   &  x    &   x   &   x   &   x   &   x   &  \\
  SAM(p)       &   x   &  x    &   x   &   x   &   x   &   x   &  \\
  mSAM(a)      &   x   &  x    &   x   &   x   &   x   &   x   &  \\
  mSAM(p)      &   x   &  x    &   x   &   x   &   x   &   x   &  \\ 
  M2F          & 0.829 & 0.770 & 0.941 & 0.907 & 0.273 & 0.183 &  \\
  \midrule
  \toprule
  \multicolumn{1}{c}{\textbf{}} &
  \multicolumn{2}{c}{\textbf{Endo}} &
  \multicolumn{2}{c}{\textbf{Golgi}} &
  \multicolumn{2}{c}{\textbf{Average}} &
\\
  \cmidrule(lr){2-3}
  \cmidrule(lr){4-5}
  \cmidrule(lr){6-7}

  & \multicolumn{1}{c}{Dice$\uparrow$} & \multicolumn{1}{c}{IoU$\uparrow$}
  & \multicolumn{1}{c}{Dice$\uparrow$} & \multicolumn{1}{c}{IoU$\uparrow$}
  & \multicolumn{1}{c}{Dice$\uparrow$} & \multicolumn{1}{c}{IoU$\uparrow$}
 \\
  \midrule
  Unet         & 0.216 & 0.155 & 0.337 & 0.224 & 0.605 & 0.522 &  \\
  SAM(a)       &   x   &  x    &   x   &   x   & 0.504 & 0.337  \\
  SAM(p)       &   x   &  x    &   x   &   x   & 0.524 & 0.355  \\
  mSAM(a)      &   x   &  x    &   x   &   x   & 0.551 & 0.380  \\
  mSAM(p)      &   x   &  x    &   x   &   x   & 0.584 & 0.412   \\ 
  M2F          & 0.537 & 0.427 & 0.408 & 0.282 & 0.672 & 0.595 &  \\
  
  \midrule
  \bottomrule
  \end{tabular}
  \end{threeparttable}
\caption{Instance segmentation performance on the homogeneous subset of the benchmark.}
    \label{tab:homogeneous}
\end{table}

\begin{table}[t]
  \centering
  \begin{threeparttable}
\setlength{\tabcolsep}{4pt}
  \begin{tabular}{
    l
    S[table-format=1.3] S[table-format=1.3]
    S[table-format=1.3] S[table-format=1.3]
    S[table-format=1.3] S[table-format=1.3]
    S[table-format=1.3] S[table-format=1.3]
    S[table-format=1.3] S[table-format=1.3]
  }
  \toprule
  \multicolumn{1}{c}{\textbf{}} &
  \multicolumn{2}{c}{\textbf{Mito}} &
  \multicolumn{2}{c}{\textbf{Nucleus}} &
  \multicolumn{2}{c}{\textbf{ER}} &
\\
  \cmidrule(lr){2-3}
  \cmidrule(lr){4-5}
  \cmidrule(lr){6-7}

  & \multicolumn{1}{c}{Dice$\uparrow$} & \multicolumn{1}{c}{IoU$\uparrow$}
  & \multicolumn{1}{c}{Dice$\uparrow$} & \multicolumn{1}{c}{IoU$\uparrow$}
  & \multicolumn{1}{c}{Dice$\uparrow$} & \multicolumn{1}{c}{IoU$\uparrow$}
 \\
  \midrule
Unet      & 0.650 & 0.481 & 0.840 & 0.724 & 0.180 & 0.099  \\
  M2F          & 0.825 & 0.702 & 0.940 & 0.887 & 0.255 & 0.145  \\
  \midrule
  \toprule
  \multicolumn{1}{c}{\textbf{}} &
  \multicolumn{2}{c}{\textbf{Endo}} &
  \multicolumn{2}{c}{\textbf{Golgi}} &
  \multicolumn{2}{c}{\textbf{Average}} &
\\
  \cmidrule(lr){2-3}
  \cmidrule(lr){4-5}
  \cmidrule(lr){6-7}

  & \multicolumn{1}{c}{Dice$\uparrow$} & \multicolumn{1}{c}{IoU$\uparrow$}
  & \multicolumn{1}{c}{Dice$\uparrow$} & \multicolumn{1}{c}{IoU$\uparrow$}
  & \multicolumn{1}{c}{Dice$\uparrow$} & \multicolumn{1}{c}{IoU$\uparrow$}
 \\
  \midrule
  Unet         & 0.210 & 0.117 & 0.300 & 0.176 & 0.436 & 0.320  \\ 
  M2F          & 0.530 & 0.360 & 0.390 & 0.242 & 0.588 & 0.467  \\
  
  \midrule
  \bottomrule
  \end{tabular}
  \end{threeparttable}
\caption{Instance segmentation performance on the full in-the-wild benchmark, reported as per-organelle Dice and IoU.}
    \label{tab:wild}

\end{table}
\section{Experiments}
 \label{sec:pagestyle}
 
 \subsection{Implementation Details}
 \label{ssec:impl_details}
 
 We trained models using the Adam optimizer (lr=0.005) for 9000 iterations (batch size 64) on four A100 GPUs. To address data heterogeneity, we applied Scale Alignment to a target resolution of 8.0 nm/pixel. However, this strategy still required using a patch-based approach 512x512 random crops during training and evaluation, which loses global context.

 \subsection{Main Results}
 \label{ssec:main_results}
 
 We evaluated all models on a homogeneous subset (Table \ref{tab:homogeneous}) and the full "in-the-wild" benchmark (Table \ref{tab:wild}). The results quantify two distinct challenges.
 
 \noindent\textbf{1. Generalization Gap from Heterogeneity.}
 A direct comparison of Table \ref{tab:homogeneous} (homogeneous) and Table \ref{tab:wild} (full benchmark) reveals a significant performance drop (e.g., M2F Average Dice from 0.672 to 0.588). This gap is driven by data heterogeneity. Despite Scale Alignment, models struggle to generalize across structural variations from different native resolutions and imaging conditions.
 
 \noindent\textbf{2. The Fundamental Mismatch: Local vs. Global.}
 More importantly, Table \ref{tab:wild} illustrates a fundamental mismatch via a stark performance dichotomy. Models perform adequately on localized organelles (Mito: 0.825 Dice, Nucleus: 0.940 Dice) but catastrophically fail on global networks (ER: 0.255 Dice, Golgi: 0.390 Dice).
 
 This failure is not due to poor training (as the Mito/Nucleus results confirm). It is the direct consequence of the patch-based methodology (Sec 4.1). Models that only see 512x512 fragments are physically incapable of resolving the global instance topology of the ER. This quantifies the failure of patch-based methods on "in-the-wild" data. U-Net's localized receptive field (ER: 0.180) performs worst, while M2F's in-patch attention (ER: 0.255) is still insufficient to resolve the true global context.
\section{Conclusion}
\label{sec:conclusion}

We introduced a large-scale benchmark for multi-organelle instance segmentation "in the wild." This benchmark highlights a fundamental mismatch between the large-scale, heterogeneous nature of real-world EM data and the local, patch-based design of current SOTA architectures. We addressed resolution heterogeneity via a Scale Alignment strategy, yet our results quantify the catastrophic failure of standard patch-based methodologies on global, network-like organelles such as the ER. This failure stems from the loss of global context during patching and the inability of models to generalize across inherent data heterogeneity arising from different native resolutions. This benchmark underscores the urgent need for new methodologies capable of bridging this fundamental mismatch.



\bibliographystyle{IEEEbib}
\bibliography{strings,refs}

\end{document}